\documentclass{article}
\usepackage{acra}
\usepackage{siunitx}
\usepackage{graphicx}
\usepackage{multirow}
\usepackage{adjustbox}
\usepackage{amsmath}
\usepackage{url}
\usepackage{hyperref}
\usepackage{algorithm}
\usepackage{algpseudocode}
\usepackage{multirow}
\usepackage{array}
\usepackage{enumitem}
\usepackage{algorithm} 
\usepackage{algpseudocode} 
\usepackage{adjustbox}
\usepackage{amsmath}
\usepackage{hyperref}
\usepackage{siunitx}
\usepackage{amssymb}
\usepackage{booktabs}
\usepackage{siunitx}
\usepackage{setspace}
\usepackage{array}
\usepackage[table,xcdraw]{xcolor}
\usepackage{xcolor}  
\usepackage{subcaption} 
\definecolor{custompurple}{HTML}{800080}  
\definecolor{customblue}{HTML}{0FF0FC}  
\definecolor{customgreen}{HTML}{008000}  
\definecolor{dodgerblue}{HTML}{4169E1}

\title{Reward Prediction Error Prioritisation in Experience Replay: The RPE-PER Method}

\author{
    Hoda Yamani$^{*}$, Yuning Xing, Lee Violet C. Ong, Bruce A. MacDonald, Henry Williams$^{**}$\\
    Centre for Automation and Robotic Engineering Science\\
    The University of Auckland, NZ\\
    \texttt{hya650@aucklanduni.ac.nz$^*$, henry.williams@auckland.ac.nz$^{**}$} \\
}

\begin{document}

\maketitle
\begin{abstract}
Reinforcement Learning algorithms aim to learn optimal control strategies through iterative interactions with an environment. A critical element in this process is the experience replay buffer, which stores past experiences, allowing the algorithm to learn from a diverse range of interactions rather than just the most recent ones. This buffer is especially essential in dynamic environments with limited experiences. However, efficiently selecting high-value experiences to accelerate training remains a challenge. Drawing inspiration from the role of reward prediction errors (RPEs) in biological systems, where they are essential for adaptive behaviour and learning, we introduce Reward Predictive Error Prioritised Experience Replay (RPE-PER). This novel approach prioritises experiences in the buffer based on RPEs. Our method employs a critic network, EMCN, that predicts rewards in addition to the Q-values produced by standard critic networks. The discrepancy between these predicted and actual rewards is computed as RPE and utilised as a signal for experience prioritisation. Experimental evaluations across various continuous control tasks demonstrate RPE-PER's effectiveness in enhancing the learning speed and performance of off-policy actor-critic algorithms compared to baseline approaches.
\end{abstract}

\section{Introduction}
     Reinforcement Learning (RL) is a powerful method for robotic control, allowing robots to learn in complex environments without explicit programming \cite{ibarz2021train}. However, RL agent's requirement for extensive interaction with the environment to develop a reliable control policy presents challenges for real-world robotic applications. Replay buffers play a crucial role in mitigating this limitation and improving its efficiency \cite{lin1992self}. These buffers store past experiences, letting the RL agent learn from and reuse previous interactions \cite{mnih2015human}. Nevertheless, in traditional RL, experiences are often randomly sampled from the replay buffer, leading to inefficient learning, especially when specific experiences hold more valuable insights than others.

    Prioritised Experience Replay (PER) \cite{schaul2015prioritized} was introduced to enhance the efficiency of the replay buffer by prioritising more informative experiences. The core idea behind PER is to improve sampling efficiency by assigning higher probabilities to experiences with greater temporal difference (TD) errors, a method that has been effective in discrete environments. However, when applied to continuous action domains, relying solely on TD error for prioritisation can lead to sampling inefficiencies and reduced performance, sometimes performing worse than random sampling \cite{saglam2022actor,oh2021model,fujimoto2018addressing}. As a result, several approaches have been explored to improve PER's performance in continuous environments. These approaches involve prioritising transitions with either higher or lower TD errors \cite{saglam2022actor}, incorporating external characteristics apart from neural networks \cite{oh2021model}, or even employing separate neural networks to estimate the prioritised scoring \cite{zha2019experience}.
    
    \begin{figure*}[t]
    \centering
    \includegraphics[width=0.7\linewidth]{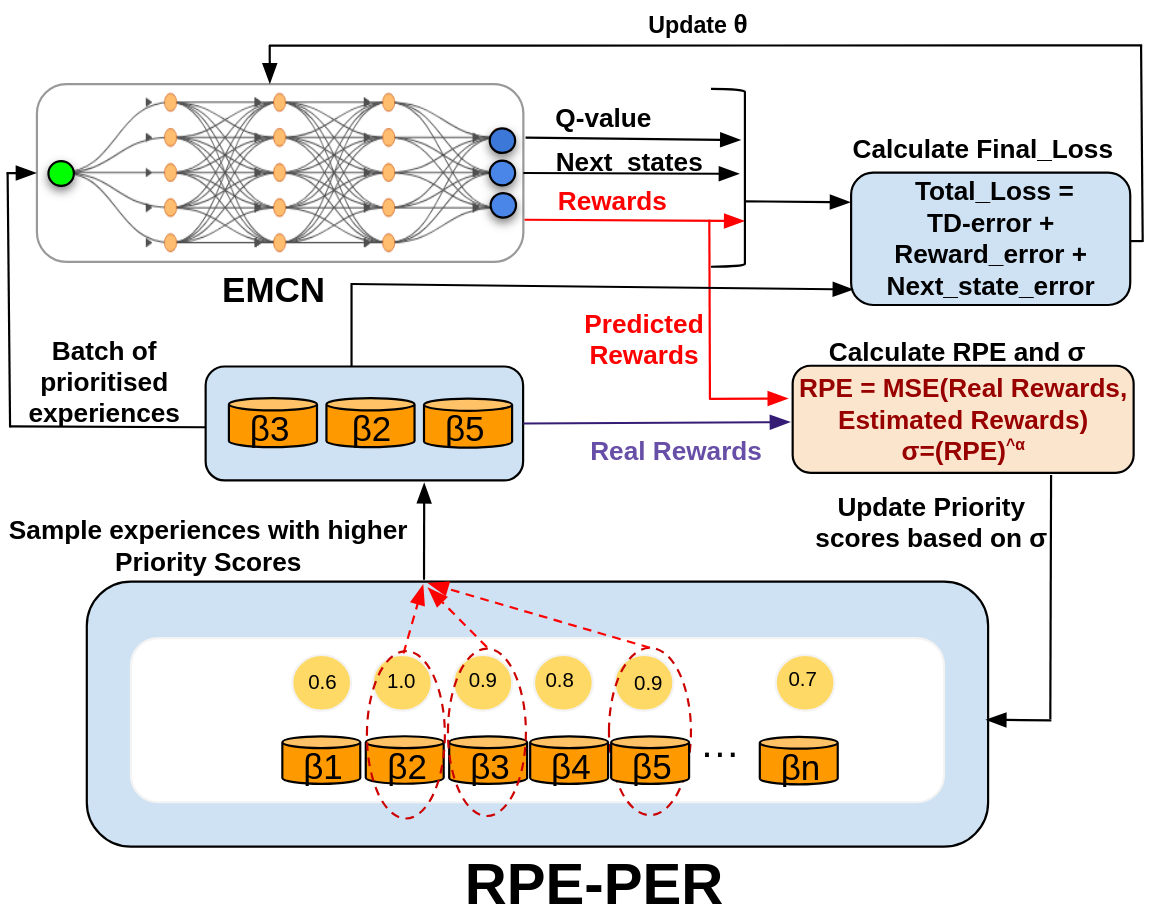}
    \caption[A high-level representation of RPE-PER and EMCN.]{A high-level representation of the RPE-PER framework and EMCN critic network. RPE is determined by computing the difference between the actual rewards stored in the buffer and the predicted rewards generated by EMCN, which is then used as a score for prioritisation experiences in the buffer.}
    \label{fig:RPE-PER}
    \end{figure*}

    Reward Prediction Error (RPE) is a foundational concept in biological systems and RL, which plays a crucial role in how organisms and artificial agents prioritise specific memories \cite{schultz2017reward},\cite{braun2018retroactive}. In the brain, particularly within the dopaminergic system, the reward prediction error hypothesis suggests that dopamine neurons encode the difference between expected and actual rewards \cite{lerner2021dopamine}. This signal is essential for adjusting behaviour, as it strengthens the memory of unexpectedly rewarding events, making them more likely to be recalled in the future \cite{rouhani2024building}. This adaptive mechanism is vital for survival, driving organisms to seek rewarding experiences while avoiding detrimental ones. In RL, RPE functions similarly by guiding the agent's learning process. The discrepancy between predicted and actual rewards informs the agent about the accuracy of its predictions, enabling it to refine its policies and improve decision-making over time \cite{simmons2019reward}. By continuously adjusting based on reward error, the agent can better align its actions with optimal outcomes, leading to more effective and efficient learning\cite{zhou2022continuously}. This alignment of biological principles with RL highlights the importance of RPE as a fundamental driver of learning and adaptation in both natural and artificial contexts. 

    This paper introduces a novel prioritisation strategy called "Reward Prediction Error Prioritised Experience Replay" (RPE-PER). This strategy uses the discrepancy between the predicted and actual rewards observed in the environment to select more valuable experiences from the buffer for training. The predicted rewards are generated by a novel critic network named the ``Enhanced Model Critic Network" (EMCN), inspired by the Model Augmented critic network \cite{oh2021model}. EMCN is an extended critic network that predicts  Q-values, next-state and rewards to model the environment effectively. The architecture of RPE-PER is shown in Fig. \ref{fig:RPE-PER}.
  
\begin{figure}[t]
        \centering
            \includegraphics[width=1.0\linewidth]{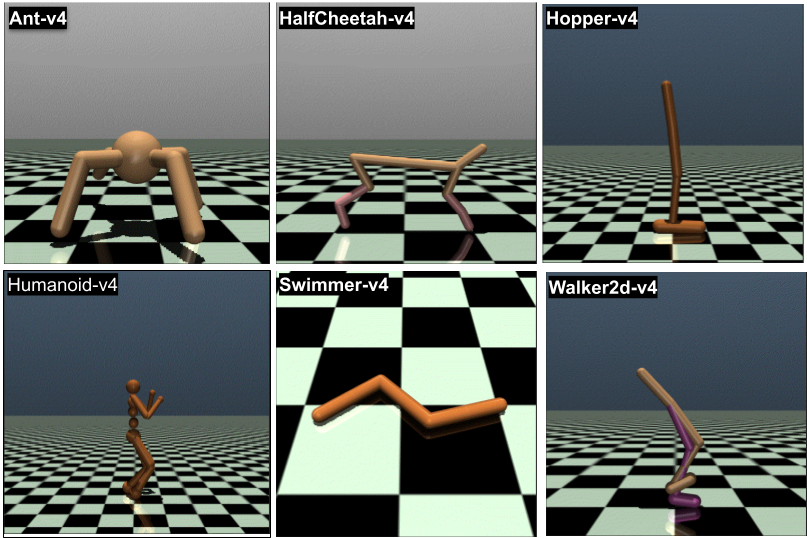}
            \caption{Six MuJoCo Tasks with Varying Complexity Levels Used in Our Experimental Evaluation
            }
            \label{fig:Mujoco-6Envs}
    \end{figure}

    By focusing on these reward prediction errors, RPE-PER prioritises experiences that are more likely to offer valuable insights into the learning process. This approach enhances sampling efficiency in continuous action domains and aligns with biological principles of memory prioritisation, making it a robust and effective method for improving RL performance. Our evaluation reveals that RPE-PER offers notable enhancements over several state-of-the-art prioritisation buffer methods and can improve performance on a variety of challenging MuJoCo continuous control benchmarks \cite{todorov2012mujoco} (see Fig. \ref{fig:Mujoco-6Envs}). RPE-PER's ability to integrate seamlessly with advanced off-policy reinforcement learning frameworks, such as TD3 \cite{fujimoto2018addressing} and SAC \cite{haarnoja2018soft}, highlights its practical utility and versatility. For reproducibility and further exploration, our open-source implementation is available on GitHub.\footnote{\url{https://github.com/UoA-CARES/RPE-PER}}.

The paper’s contributions can be summarised as follows:
\vspace{-0.5em}
\begin{enumerate}
    \item Introduce a novel prioritisation strategy integrating Reward Prediction Error from biological concepts with experience replay to enhance learning efficiency.
    \item Use EMCN, a more comprehensive critic network for reward prediction that also predicts the model of the environment to achieve a better estimation of rewards.
    \item Validate the effectiveness of RPE-PER in complex MuJoCo environments using SAC and TD3 algorithms, focusing on improved sampling efficiency and enhanced learning outcomes.
\end{enumerate}
    
\section{Related Work}
     Initially proposed by Schaul et al. \cite{schaul2015prioritized}, Prioritised Experience Replay (PER) is a pivotal technique in deep reinforcement learning due to its ability to accelerate learning by replaying more informative transitions. By prioritising experiences based on their TD error, PER increases the likelihood that transitions with higher errors will be replayed, thereby enhancing the learning efficiency of deep Q-networks (DQN) \cite{mnih2015human}. Despite its success in discrete action domains, PER's efficacy in continuous control environments has been questioned, particularly regarding its ability to prioritise critical experiences consistently.
    
    To address some of these limitations, Loss-Adjusted Prioritised (LAP) \cite{fujimoto2020equivalence} improves upon traditional PER by optimising the sampling distribution based on the loss function rather than relying solely on TD error. Although LAP offers a more targeted approach to experience replay and can enhance learning efficiency, it still depends on the assumption that the loss function alone is sufficient for effective sampling, which may not hold in complex environments with sparse or misleading reward signals. Building on this, Saglam et al. introduced Loss-Adjusted Approximate Actor Prioritised Experience Replay (LA3P) \cite{saglam2022actor}. LA3P addresses PER's shortcomings in actor-critic architectures by prioritising low TD error transitions for actor training and high TD error transitions for critic training while incorporating random sampling for balance. Nonetheless, it relies on TD error for prioritisation, which can be problematic due to its overemphasis on immediate rewards, high variance, and limited effectiveness in capturing long-term impacts.
    
    In contrast, Oh et al. \cite{oh2021model} critique TD error-based sampling for its inefficiencies stemming from inaccurate Q-value estimates due to deep function approximators and bootstrapping. They propose MaCN, a novel critic network inspired by augmented rewards for exploration \cite{pathak2017curiosity}, \cite{shelhamer2016loss}, which predicts additional features like rewards and transitions alongside Q-values. They also developed MaPER, a model-augmented replay buffer that uses these features to calculate prioritisation scores. While MaPER shows promise for enhancing the learning process, there remains room for improvement in efficiency.

    Prior research has explored various scoring methods for prioritising past experiences in RL, often relying on TD error, which yields limited improvements in challenging continuous environments. Therefore, there is a need to investigate alternative scaling mechanisms for experience prioritisation. Our approach, Reward Prediction Error Prioritised Experience Replay (RPE-PER), draws inspiration from biological learning mechanisms \cite{lerner2021dopamine}. RPE-PER aims to enhance experience selection that aligns more closely with our brain's biological learning mechanisms by focusing on the discrepancies between predicted and actual rewards \cite{rouhani2024building}. This approach is expected to improve learning outcomes in complex environments where traditional TD error signals may be inadequate. Our empirical studies compare RPE-PER against methods such as random sampling, PER, LAP, LA3P, and MaPER to evaluate its effectiveness and potential benefits.

\section{\textbf{Methodology}}\label{sec:methods}

This section presents the methodology for developing the RPE-PER framework. It commences with a discussion of foundational concepts in RL and PER. Subsequently, the methodology delineates the role of the EMCN critic network in reward prediction. The process then transitions to the calculation of the RPE, defined as the discrepancy between the predicted rewards and the actual rewards stored in the replay buffer. This RPE serves as a critical criterion for prioritizing experiences during training, enabling more efficient learning.

\subsection{\textbf{Preliminaries}}

\subsubsection{\textbf{Reinforcement Learning}}
Deep Reinforcement Learning integrates RL algorithms with deep neural networks to learn policies or value functions from complex, high-dimensional inputs. Within this context, the RL framework is formalised as a finite Markov Decision Process (MDP), denoted as a 5-tuple $(S, A, R, P, \gamma)$, where $S$ is the state space, $A$ is the action space, $R$ represents the reward function (deterministic or stochastic), $P$ characterise the environment dynamics as the state transition probability distribution and $\gamma$ is the discount factor within the range $[0, 1)$.

An agent's objective in this MDP framework is to learn an optimal policy $\pi: S \rightarrow A$ that maximises the expected cumulative reward. This is typically achieved by estimating the action-value function $Q_\pi(s, a)$, defined as:

\begin{equation}\label{eq:1}
    Q_\pi(s, a) = \mathbb{E}_{s', r} \left[ r + \gamma Q_\pi(s', \pi(s')) \mid s, a \right]
\end{equation}

\subsubsection{\textbf{Prioritised Experience Replay}}
To select valuable past experiences for learning, PER is employed \cite{schaul2015prioritized}. Instead of random sampling, PER assigns a priority to each transition tuple $(s_t, a_t, r_t, s_{t+1})$ stored in the replay buffer $B$. The priority of each transition is calculated based on the TD error, which quantifies the difference between the expected and the actual Q-values in \eqref{eq:1} \cite{brittain2019prioritized}. The TD error for each transition $t$ is calculated as:

\begin{equation}\label{eq:2}
    \delta_t = \left| r_t + \gamma \max_{a'} Q(s_{t+1}, a'; \theta^{-}) - Q(s_t, a_t; \theta) \right|
\end{equation}

Here, $\theta$ represents the parameters of the current Q-network, and $\theta^{-}$ corresponds to the parameters of the target Q-network, which are periodically updated to stabilize learning.

The priority $p_i$ for each transition $i$ is computed by:

\begin{equation}\label{eq:3}
    p_i = |\delta_i| + \epsilon
\end{equation}

Where $\epsilon$ is a small positive constant added to ensure that all transitions have a non-zero priority, preventing them from being entirely ignored.

The probability $P(i)$ of sampling a particular transition $i$ from the replay buffer is then proportional to its priority $p_i^\alpha$:

\begin{equation}\label{eq:4}
    P(i) = \frac{p_i^\alpha}{\sum_{k} p_k^\alpha}
\end{equation}

Where $\alpha$ is a hyper-parameter that controls the degree of prioritisation. A higher $\alpha$ emphasises transitions with larger TD errors, meaning they are more likely to be sampled during training.

The replay buffer is updated by inserting new transition tuples as follows:

\begin{equation}\label{eq:5}
    B \leftarrow B \cup \{(s_t, a_t, r_t, s_{t+1})\}
\end{equation}

This prioritisation mechanism ensures that transitions with higher TD errors, which indicate a larger discrepancy between expected and actual outcomes, are sampled more frequently. This focus on high-priority transitions accelerates learning by allowing the agent to correct its predictions and improve its policy more efficiently.

\subsection{\textbf{Enhanced Model Critic Network (EMCN)}}\label{sec:emcn-network}

We introduce the EMCN, an advanced critic network \( C_{\theta}(s, a) \) that not only estimates Q-values \( Q_{\theta}(s, a) \) but also predicts rewards \( {R}_{\theta}(s, a) \) and next-state transitions \( {T}_{\theta}(s, a) \). By predicting rewards and next states for a given state-action pair, our critic network effectively constructs a model of the environment within its predictions \cite{januszewski2022model}. Consequently, EMCN, by having an environmental model that provides insights into potential outcomes, can enhance exploration and improve learning efficiency by reducing the need for extensive real-world interactions through simulated experiences \cite{ha2018world}. The critic network is represented as:

\begin{equation}\label{eq:7}
    C_{\theta} = \left( Q_{\theta}(s, a), R_{\theta}(s, a), T_{\theta}(s, a) \right)
\end{equation}

Our EMCN is inspired by MaCN introduced in \cite{oh2021model}. A critical distinction between EMCN and MaCN is that while MaCN uses estimated rewards \( \hat{R}_{\theta} \) to compute Q-values in order to drive exploration \cite{hessel2018rainbow}, EMCN relies on actual rewards. Our approach avoids biases that may arise from using estimated rewards. 

The prediction error for Q-values \( \delta_t Q_{\theta} \) is computed using real rewards:

\begin{equation}\label{eq:8}
    \delta_t Q_{\theta} = r_t + \gamma \mathbb{E}_{a' \sim \pi_{\Theta}(\cdot|s_{t+1})} \left[ Q_{\theta}(s_{t+1}, a') \right] - Q_{\theta}(s_t, a_t)
\end{equation}

The Reward Prediction Error \( \delta_t {R}_{\theta} \) and transition error \( \delta_t {T}_{\theta} \) are defined as:

\begin{equation}\label{eq:9}
    \delta_t R_{\theta} = R_{\theta}(s_t, a_t) - r_t
\end{equation}

\begin{equation}\label{eq:10}
    \delta_t T_{\theta} = T_{\theta}(s_t, a_t) - s_{t+1}
\end{equation}

The loss functions for the reward model \( \hat{R}_{\theta} \), transition model \( \hat{T}_{\theta} \), and Q-value function \( Q_{\theta} \) are given by:

\begin{equation}\label{eq:11}
\begin{aligned}
    \mathcal{L}_{R_{\theta}} &= \mathbb{E}_{(s_t, a_t, r_t, s_{t+1}) \sim B} \left[ \| \delta_t R_{\theta} \|_{\text{MSE}} \right] \\
    \mathcal{L}_{T_{\theta}} &= \mathbb{E}_{(s_t, a_t, r_t, s_{t+1}) \sim B} \left[ \| \delta_t T_{\theta} \|_{\text{MSE}} \right] \\
    \mathcal{L}_{Q_{\theta}} &= \mathbb{E}_{(s_t, a_t, r_t, s_{t+1}) \sim B} \left[ \| \delta_t Q_{\theta} \|_{\text{MSE}} \right]
\end{aligned}
\end{equation}

The overall loss function for EMCN is:

\begin{equation}\label{eq:12}
    \mathcal{L}_{C_{\theta}} = \xi_1 \mathcal{L}_{Q_{\theta}} + \xi_2 \mathcal{L}_{\hat{R}_{\theta}} + \xi_3 \mathcal{L}_{\hat{T}_{\theta}}
\end{equation}

where \( \xi_1 \), \( \xi_2 \), and \( \xi_3 \) are hyperparameters balancing the contributions of each component. Increasing \( \xi_2 \) improves performance, indicating the importance of accurate reward prediction.

\subsection{Reward Prediction Error (RPE) Calculation}

Our approach prioritizes experiences in the replay buffer based on RPE. The $\text{RPE}i$ measures the discrepancy between the actual reward received during the agent’s interaction with the environment, denoted as $r_i$, and the reward predicted by the critic network for the $i$-th experience, $R{\theta}(s_i, a_i)$, which is the predicted reward for state $s_i$ and action $a_i$, as defined in equation \eqref{eq:7}. The RPE is calculated as:

\begin{equation}\label{eq:13} \text{RPE}i = \left| R{\theta}(s_i, a_i) - r_i \right|_{\text{MSE}} \end{equation}

where $|\cdot|_{\text{MSE}}$ represents the Mean Squared Error norm. To compute priority, equation \eqref{eq:3} in the PER algorithm is replaced with:

\begin{equation}\label{eq:14} p_i = |\text{RPE}_i| + \epsilon \end{equation}

Experiences are then selected based on this priority using the probability formula in equation \eqref{eq:4}, with higher RPE values increasing the likelihood of selection for training.

The pseudo-code for RPE-PER is provided in algorithm \ref{alg:RPE-PER}. 


\begin{algorithm}[!h]
\begin{algorithmic}[1]
    \State \textbf{Initialise} critic parameters $\theta$, actor parameters $\Theta$, replay buffer $B \leftarrow \emptyset$, priority set $P_B \leftarrow \emptyset$, and batch size $m$.
    \For{each time step $t$}
        \State Choose action $a_t$ using $\pi_{\Theta}(s_t)$.
        \State Execute $a_t$, observe reward $r_t$ and next state $s_{t+1}$.
        \State Compute Reward Prediction Error (RPE):
        \[
        \text{RPE} = \left| R_{\theta}(s_t, a_t) - r_t \right|,
        \]
        where $R_{\theta}(s_t, a_t)$ is the predicted reward.
        \State Update replay buffer: $B \leftarrow B \cup \{ (s_t, a_t, r_t, s_{t+1}) \}$.
        \State Update priority set: $P_B \leftarrow P_B \cup \{ \text{RPE} \}$.
        \For{each update step}
            \State Sample mini-batch $\{\tau_i\}_{i \in I}$ using:
            \[
            I \sim \frac{(\text{RPE}_i)^\alpha}{\sum_j (\text{RPE}_j)^\alpha}; \quad |I| = m
            \]
            \State Update critic parameters $\theta$:
            \[
            \mathcal{L}_{\theta} = \xi_1 \mathcal{L}_{\mathcal{Q}} + \xi_2 \mathcal{L}_{R} + \xi_3 \mathcal{L}_{T}
            \]
            \State Update actor parameters $\Theta$ using selected mini-batch of experiences.
            \State Recalculate priorities:
            \[
            p_i = (\text{RPE}_i)^\alpha
            \]
            \State Update priority set $P_B$.
        \EndFor
    \EndFor
\end{algorithmic}
\caption{Reward Prediction Error Prioritised Experience Replay (RPE-PER) for Actor-Critic}
\label{alg:RPE-PER}
\end{algorithm}

\section{\textbf{Experiments}}

 \subsection{\textbf{Experimental Setup}}

    We evaluated our approach on a standardised set of continuous control tasks from the MuJoCo environment \cite{todorov2012mujoco}, which includes six distinct environments: Ant, Humanoid, HalfCheetah, Hopper, Swimmer, and Walker2d (Fig. \ref{fig:Mujoco-6Envs}). These environments, which vary in complexity, are widely recognised benchmarks for assessing the performance of RL algorithms. The tasks are accessed via the OpenAI Gym interface \cite{brockman2016openai}, ensuring seamless integration with our testing framework.

    To assess the effectiveness of RPE-PER, we compared its performance against several prominent baselines in prioritised experience replay: PER \cite{schaul2015prioritized}, LAP \cite{fujimoto2020equivalence}, LA3P \cite{saglam2022actor}, and MaPER \cite{oh2021model}, as well as random-sampling buffers\cite{mnih2015human}. Each of these replay strategies was integrated with state-of-the-art RL algorithms TD3 \cite{fujimoto2018addressing} and SAC \cite{haarnoja2018soft} for a comprehensive evaluation.

    For implementation, we used a batch size of 256 and set the actor and critic learning rates to 3e-4, with two network hidden layers consisting of 256 neurons based on \cite{fujimoto2018addressing}, \cite{haarnoja2018soft}. Our buffer RPE-PER, was configured with $\beta$ set to 0.4 and $\alpha$ to 0.7, as described in \cite{oh2021model}. Baselines were implemented with their original hyper-parameters, confirming their optimal performance when precisely tuned. We trained RPE-PER and the baselines for one million steps, using ten seeds for robustness. Performance was evaluated using average cumulative rewards per step, with evaluations every 10,000 steps to generate learning curves.






        \begin{figure*}[htbp]
    \centering
    \begin{subfigure}[b]{0.48\textwidth}
        \includegraphics[width=\textwidth]{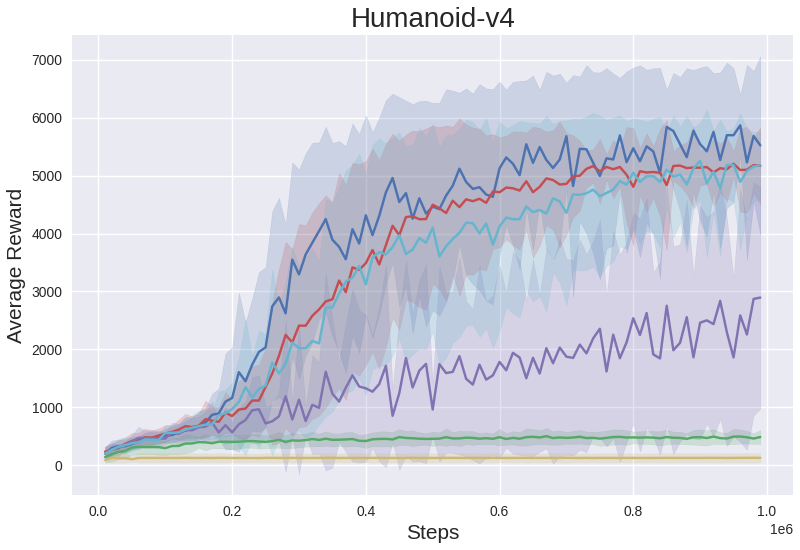}
        \caption{}
    \end{subfigure}
    \hfill
    \begin{subfigure}[b]{0.48\textwidth}
        \includegraphics[width=\textwidth]{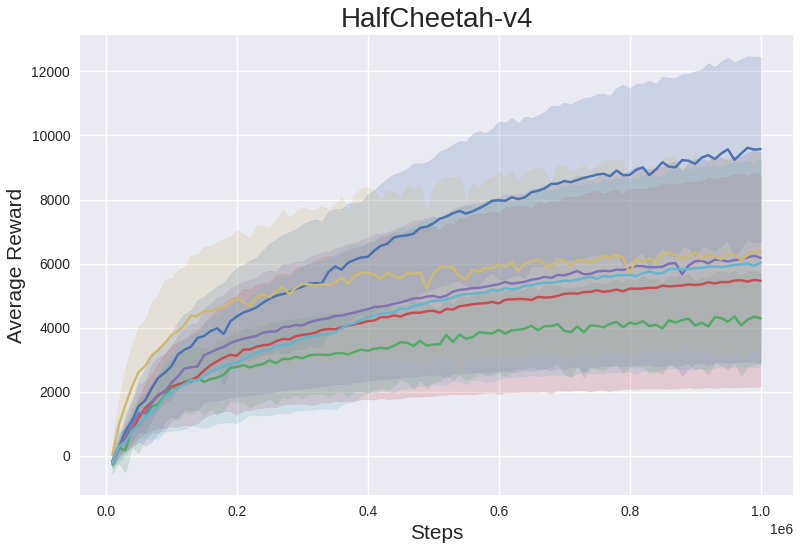}
        \caption{}
    \end{subfigure}
    \vspace{5pt} 
    \begin{subfigure}[b]{0.48\textwidth}
        \includegraphics[width=\textwidth]{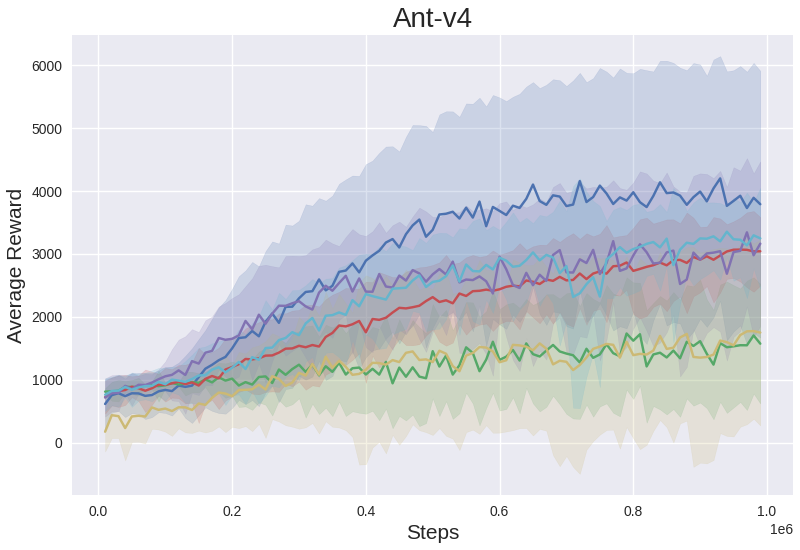}
        \caption{}
    \end{subfigure}
    \hfill
    \begin{subfigure}[b]{0.48\textwidth}
        \includegraphics[width=\textwidth]{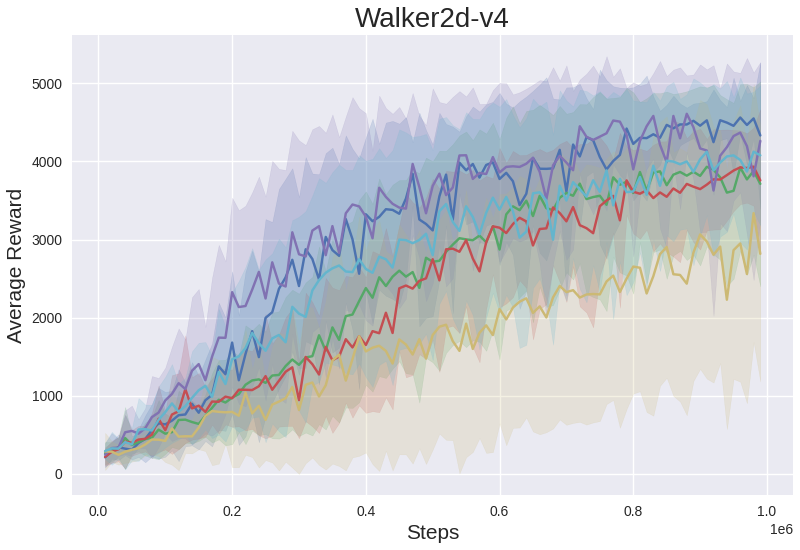}
        \caption{}
    \end{subfigure}
    \begin{subfigure}[b]{0.48\textwidth}
        \includegraphics[width=\textwidth]{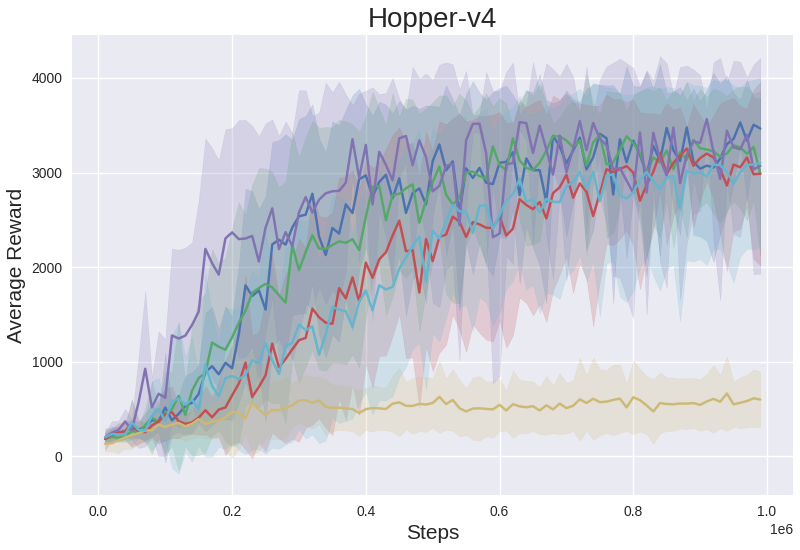}
        \caption{}
    \end{subfigure}
    \hfill
    \begin{subfigure}[b]{0.48\textwidth}
    \includegraphics[width=\textwidth]{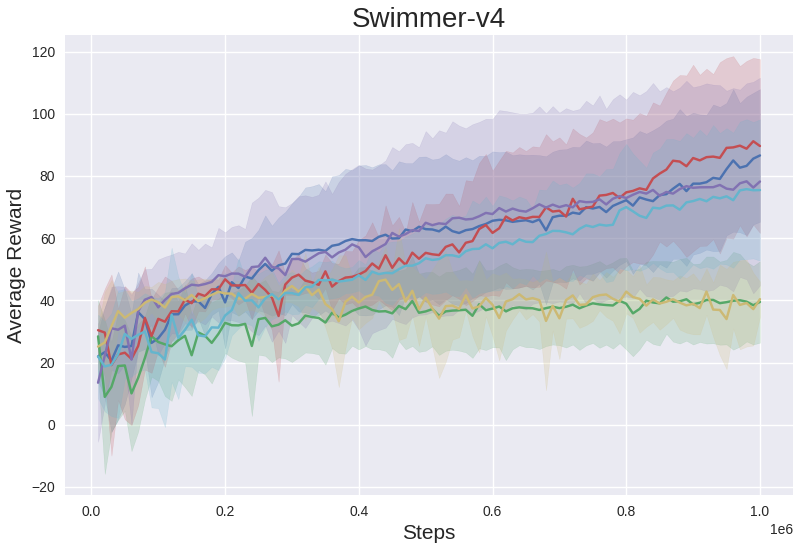}
    \caption{}
    \end{subfigure}

  
\begin{center}  
\begin{tabular}{c @{\hspace{4pt}} c @{\hspace{4pt}} c @{\hspace{4pt}} c @{\hspace{4pt}} c @{\hspace{4pt}} c}  
    \textcolor{dodgerblue}{\rule{0.05\linewidth}{0.15cm}} \textbf{\scriptsize RPE-PER} &
    \textcolor{yellow}{\rule{0.05\linewidth}{0.15cm}} \textbf{\scriptsize MAPER} &
    \textcolor{red}{\rule{0.05\linewidth}{0.15cm}} \textbf{\scriptsize PER} &
    \textcolor{customgreen}{\rule{0.05\linewidth}{0.15cm}} \textbf{\scriptsize LAP} &
    \textcolor{custompurple}{\rule{0.05\linewidth}{0.15cm}} \textbf{\scriptsize LA3P} &
    \textcolor{customblue}{\rule{0.05\linewidth}{0.15cm}} \textbf{\scriptsize Random-Sampling}
\end{tabular}
\end{center}

    \caption{Learning curves are generated in selected continuous control tasks in  MuJoCo under \textbf{TD3} policy method and using various sampling methods. Smoothing the curves for improved visual clarity is achieved by a sliding window of size 10. The solid lines represent the mean values, while the shaded regions indicate the standard deviations calculated from ten evaluations across ten runs with different random seeds.}
    \label{fig:TD3-results}
\end{figure*}
        

\begin{figure*}[htbp]
    \centering
    \begin{subfigure}[b]{0.48\textwidth}
        \includegraphics[width=\textwidth]{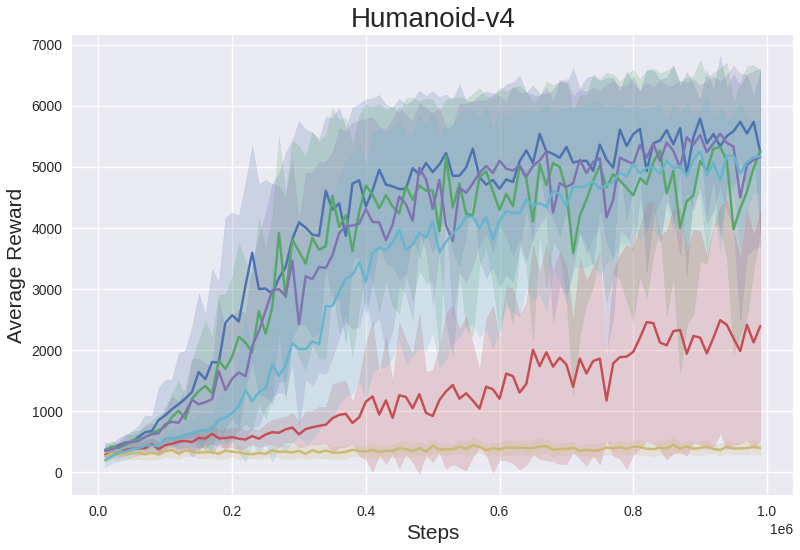}
        \caption{}
    \end{subfigure}
    \hfill
    \begin{subfigure}[b]{0.48\textwidth}
        \includegraphics[width=\textwidth]{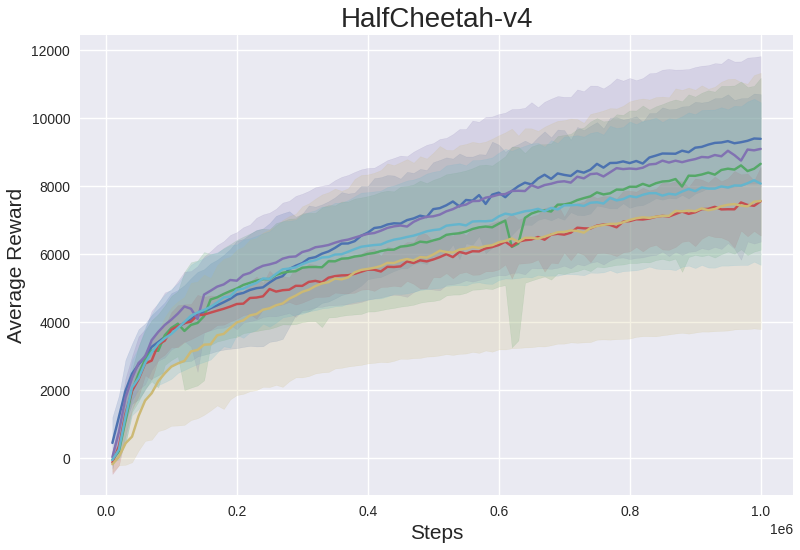}
        \caption{}
    \end{subfigure}
    \vspace{5pt} 
    \begin{subfigure}[b]{0.48\textwidth}
        \includegraphics[width=\textwidth]{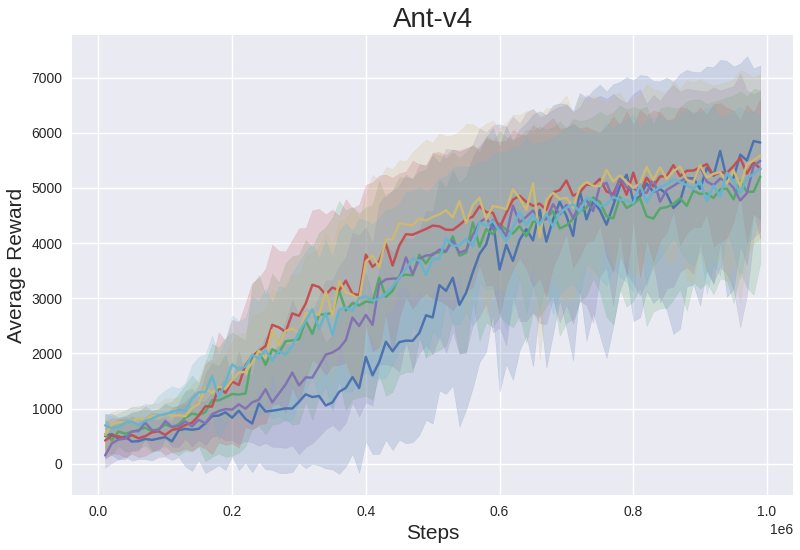}
        \caption{}
    \end{subfigure}
    \begin{subfigure}[b]{0.48\textwidth}
        \includegraphics[width=\textwidth]{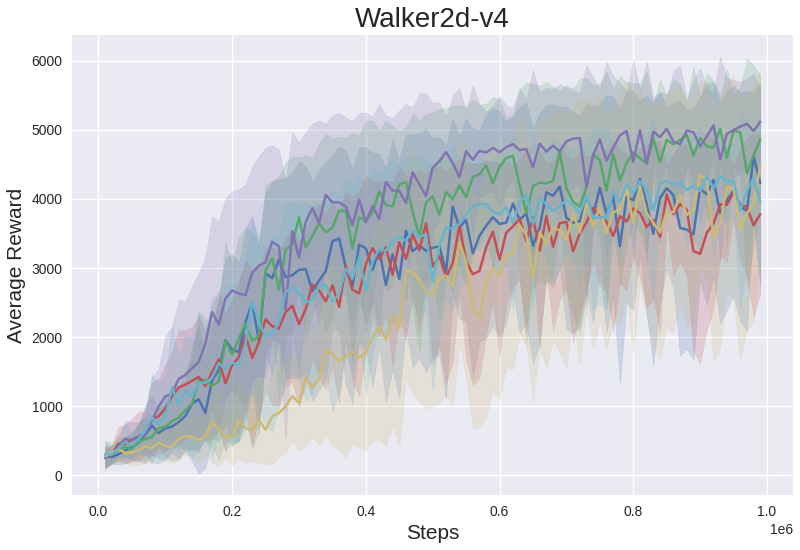}
        \caption{}
    \end{subfigure}
    \hfill
    \begin{subfigure}[b]{0.48\textwidth}
        \includegraphics[width=\textwidth]{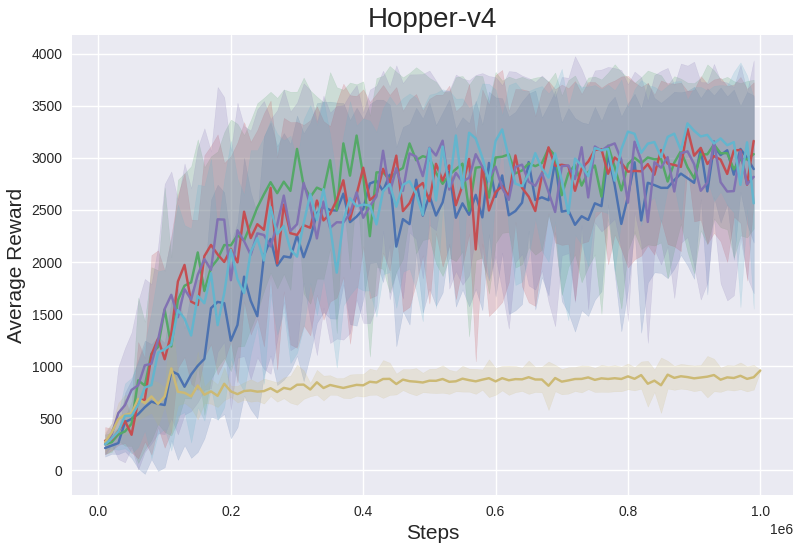}
        \caption{}
    \end{subfigure}
    \begin{subfigure}[b]{0.48\textwidth}
        \includegraphics[width=\textwidth]{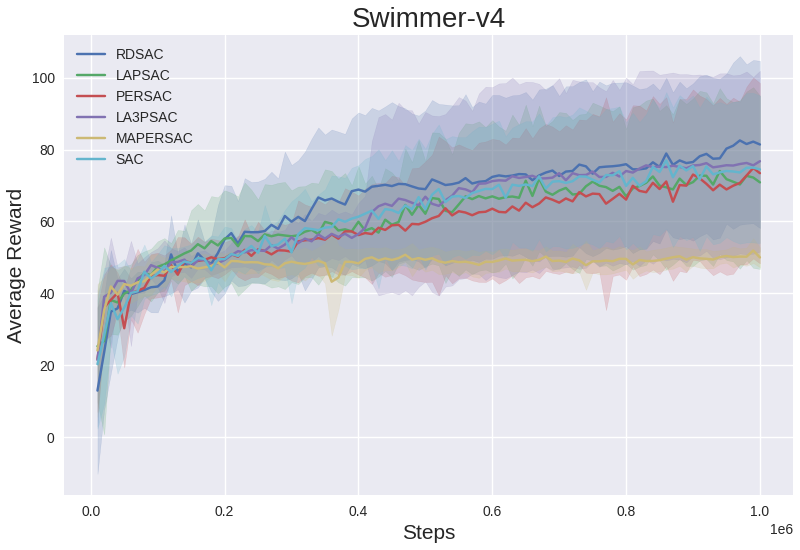}
        \caption{}
    \end{subfigure}
    \hfill

\begin{center}  
\begin{tabular}{c @{\hspace{4pt}} c @{\hspace{4pt}} c @{\hspace{4pt}} c @{\hspace{4pt}} c @{\hspace{4pt}} c}  
    \textcolor{dodgerblue}{\rule{0.05\linewidth}{0.15cm}} \textbf{\scriptsize RPE-PER} &
    \textcolor{yellow}{\rule{0.05\linewidth}{0.15cm}} \textbf{\scriptsize MAPER} &
    \textcolor{red}{\rule{0.05\linewidth}{0.15cm}} \textbf{\scriptsize PER} &
    \textcolor{customgreen}{\rule{0.05\linewidth}{0.15cm}} \textbf{\scriptsize LAP} &
    \textcolor{custompurple}{\rule{0.05\linewidth}{0.15cm}} \textbf{\scriptsize LA3P} &
    \textcolor{customblue}{\rule{0.05\linewidth}{0.15cm}} \textbf{\scriptsize Random-Sampling}
\end{tabular}
\end{center}

    \caption{Learning curves are generated in selected continuous control tasks in MuJoCo under the \textbf{SAC} policy method using various sampling methods. Smoothing the curves for improved visual clarity is achieved by a sliding window of size 10. The solid lines represent the mean values, while the shaded regions indicate the standard deviations, calculated from ten evaluations across ten runs with different random seeds.}
    \label{fig:SAC-results}
\end{figure*}
\begin{table*}[!h]
        \centering
        \caption{The average performance over the last 10 evaluation results with ± the 95\% confidence interval across 10 different seeds under the \textbf{TD3} algorithm. Bold scores indicate the best-performing result.}
        \vspace{0.5em} 
        \label{tab:performance-TD3}
        \begin{tabular}{|p{3.5cm}|p{3.5cm}|p{3.5cm}|p{3.5cm}|}
        \hline
        \textbf{Method} & \textbf{Humanoid} & \textbf{HalfCheetah} & \textbf{Ant} \\
        \hline
        \textbf{RPE-PER} & \textbf{5522.25 $\pm$1540.32} & \textbf{9572.35 $\pm$2880.03} & \textbf{3790.61 $\pm$2119.25}  \\
        PER & 5169.45$\pm$663.16 & 5467.76$\pm$3300.72 & 3043.24$\pm$542.73 \\
        MaPER & 247.10 $\pm$162.67 & 7045.82$\pm$2066.65 & 2039.67$\pm$1355.37 \\ 
        LAP & 485.16$\pm$119.91 & 4292.03$\pm$1391.53 & 1575.09$\pm$949.84 \\ 
        LA3P & 2894.06$\pm$1921.65 & 6180.07$\pm$3304.63 & 3164.64$\pm$1306.27  \\ 
        \textbf{Random-Sampling} & 5177.45$\pm$585.13 & 6036.24$\pm$3238.51 & 3248.08$\pm$802.71 \\ 
        \hline
        \end{tabular}
\end{table*}
    
    \begin{table*}[!h]
        \centering
        \label{tab:performance-TD3-2}
        \begin{tabular}{|p{3.5cm}|p{3.5cm}|p{3.5cm}|p{3.5cm}|}
        \hline
        \textbf{Method} & \textbf{Walker2d} & \textbf{Hopper} & \textbf{Swimmer} \\
        \hline
        \textbf{RPE-PER} & \textbf{4331.51$\pm$935.03} & \textbf{3464.80$\pm$322.23} & 86.61$\pm$21.34 \\ 
        PER & 3756.14$\pm$909.93 & 2986.99$\pm$968.46 & \textbf{89.64$\pm$27.97} \\
        MaPER & 2816.97$\pm$1632.46 & 541.68$\pm$335.02 & 40.39$\pm$5.64 \\ 
        LAP & 3712.27$\pm$1311.46 & 2988.96$\pm$873.84 & 39.49$\pm$13.09 \\ 
        LA3P & 4262.39$\pm$1006.90 & 3070.85$\pm$1139.10 & 78.22$\pm$33.39 \\ 
        \textbf{Random-Sampling} & 4076.80$\pm$894.46 & 3104.16$\pm$893.19 & 75.47$\pm$22.73 \\ 
        \hline
        \end{tabular}
    \end{table*}
      \begin{table*}[!h]
        \centering
        \caption{The average performance over the last 10 evaluation results with ± the 95\% confidence interval across 10 different seeds under the \textbf{SAC} algorithm. Bold scores indicate the best-performing result.}
        \vspace{0.5em} 
        \label{tab:performance-SAC}
        \begin{tabular}{|p{3.5cm}|p{3.5cm}|p{3.5cm}|p{3.5cm}|}
        \hline
        \textbf{Method} & \textbf{Humanoid} & \textbf{HalfCheetah} & \textbf{Ant} \\
        \hline
        \textbf{RPE-PER} & \textbf{5239.34 $\pm$1308.53} & \textbf{9395.45 $\pm$1313.66} & \textbf{5821.84 $\pm$1397.28}  \\
        PER & 2396.44 $\pm$1863.55 & 7558.46 $\pm$1010.69 & 5344.69$\pm$1267.14 \\
        MaPER & 400.03 $\pm$112.90 & 7567.58 $\pm$3768.51 & 5586.07$\pm$1490.59 \\ 
        LAP & 4985.30 $\pm$1677.68 & 7558.46 $\pm$1010.69 & 5203.80$\pm$1565.88 \\ 
        LA3P & 5158.66 $\pm$1448.85 & 9099.23 $\pm$2728.37 & 5488.88$\pm$1266.80  \\ 
        \textbf{Random-Sampling} & 5177.45 $\pm$585.14 & 8077.38 $\pm$2389.68 & 5356.86$\pm$949.15 \\ 
        \hline
        \end{tabular}
    \end{table*}
    
    \begin{table*}[!h]
        \centering
        \label{tab:performance-SAC1}
        \begin{tabular}{|p{3.5cm}|p{3.5cm}|p{3.5cm}|p{3.5cm}|}
        \hline
        \textbf{Method} & \textbf{Walker2d} & \textbf{Hopper} & \textbf{Swimmer} \\
        \hline
        \textbf{RPE-PER} & 4231.47$\pm$1462.85 & 2893.03$\pm$706.76 & \textbf{81.43$\pm$23.27} \\ 
        PER & 3780.41$\pm$1159.49 & \textbf{3162.61$\pm$562.59} & 73.44$\pm$25.10 \\
        MaPER  & 4486.17 $\pm$1392.27 & 959.00$\pm$7.42 & 50.00$\pm$2.73 \\ 
        LAP & 4865.81$\pm$921.12 & 2987.00$\pm$751.23 & 70.89$\pm$24.9 \\ 
        LA3P & \textbf{5119.40$\pm$485.90} & 2818.22$\pm$1117.58 & 76.77$\pm$25.17 \\ 
        \textbf{Random-Sampling} & 3962.01$\pm$1284.04 & 2564.37$\pm$1012.43 & 74.48$\pm$20.48 \\ 
        \hline
        \end{tabular}
    \end{table*}

    \subsection{\textbf{Learning Efficiency Analysis}}
    
       Learning curves, which compare the learning efficiency of RPE-PER and the baseline methods, are displayed in Fig. \ref{fig:TD3-results} under the TD3 algorithm and Fig. \ref{fig:SAC-results} under the SAC algorithm. The curves illustrate the performance across six Mujoco continuous control environments. These curves illustrate performance progression over evaluation steps, highlighting the convergence speed and distinctions between RPE-PER and the baseline methods.
       
       Furthermore, we provide the mean of the final ten total evaluation results, representing convergence points for all algorithms, in Table \ref{tab:performance-TD3} for the TD3 versions and Table \ref{tab:performance-SAC} for SAC versions. While some baselines underperformed compared to their published results due to stochasticity, different seeds, and environment updates, these variances do not affect the consistency of performance differences. Consequently, our comparative assessments adhere to the established standards for equitable deep reinforcement learning benchmarking \cite{henderson2018deep}.



\section{\textbf{Experimental Results}}





        
\subsection{\textbf{Performance Analysis and Discussion}}

    Based on our evaluation results, integrating RPE-PER into the TD3 and SAC algorithms yields performance improvements compared to baseline algorithms.

    In particular, RPE-PER combined with TD3 consistently outperforms most baseline methods across various MuJoCo environments, as illustrated in (Fig. \ref{fig:TD3-results}, plots a to e). This result confirms the effectiveness of prioritising based on Reward Prediction Error, demonstrating significant advantages in complex environments. The exception is the Swimmer task (Fig. \ref{fig:TD3-results}, plot f), where the simpler PER algorithm slightly outperforms RPE-PER. This discrepancy suggests that in straightforward, lower-dimensional tasks like Swimmer, the benefits of RPE-PER may be less pronounced, making the simpler PER algorithm more practical. Despite this, the overall superior performance of RPE-PER in more complex tasks highlights the robustness and generalisability of our approach, validating its effectiveness in a broader range of environments.
    
    Additionally, comparing RPE-PER to MaPER within both TD3 and SAC algorithms reveals that RPE-PER delivers superior performance. Despite MaPER employing more complex prioritisation scores derived from model prediction errors, RPE-PER's use of RPE proves more effective in identifying and using valuable experiences, leading to better overall performance.

    
    
    RPE-PER also shows performance enhancement when combined with the SAC algorithm, particularly in tasks such as Humanoid, HalfCheetah, Ant, and Swimmer (Fig. \ref{fig:SAC-results}, plots a to c and f) compared to baseline approaches. However, in the Walker2d and Hopper tasks (Fig. \ref{fig:SAC-results}, plots d and e), RPE-PER slightly underperforms compared to some baseline algorithms. This underperformance may be attributed to the intricate locomotion dynamics of these tasks, which demand precise coordination and balance in high-dimensional action spaces. 
    
    For instance, Walker2d requires careful coordination of two legs to maintain a stable gait, while Hopper involves balancing a single-legged robot, where any misjudgment in force application can lead to instability and falls. RPE-PER prioritises transitions where the agent's reward predictions are most inaccurate. While this can accelerate learning in many environments, it may lead to over-prioritization of highly variable or noisy transitions in tasks like Walker2d and Hopper, where precise control is crucial. This over-prioritisation could disrupt the delicate balance required for these tasks, leading to less stable learning and slightly lower performance compared to other environments. So, Traditional TD error-based algorithms, which focus on refining long-term policy stability, may perform better in these scenarios.

    The other important point is about SAC’s design, with its focus on optimising policies in continuous action spaces by augmenting rewards with an entropy term, thereby promoting exploration in states with high future entropy. This approach primarily focuses on optimizing the policy directly while maintaining a stable entropy balance. As a result, the impact of reward prediction errors may be diluted, reducing the influence of experience prioritization based on these errors. Thus, while RPE-PER provides some performance benefits in SAC, these gains are more modest compared to TD3. These mixed results suggest that while RPE-PER can improve SAC performance, its benefits are more evident in TD3, where the focus is directly on reward prediction accuracy and Q-value optimization.
    
    Further noteworthy results in the SAC algorithm results in Fig. \ref{fig:SAC-results} can be the LA3P technique. Although LA3P underperforms compared to RPE-PER in most tasks (Fig. \ref{fig:SAC-results}, plots a to c and f), it performs comparably or even outperforms RPE-PER in Walker2d and Hopper (Fig. \ref{fig:TD3-results}, plots d and e). This good performance indicates that LA3P's strategy of prioritising experiences with lower TD error for the actor and higher TD error for the critic could offer valuable insights, particularly for the SAC algorithm in tasks requiring precise balance and control.
    
    Future work for this research will investigate several avenues to enhance performance further. One approach is to explore signed RPE for actor and critic networks inspired by the results of LA3P in SAC (Fig. \ref{fig:SAC-results}). Another potential direction is to integrate an entropy-based component into the RPE-PER strategy. By incorporating the entropy of the action distribution at the time of experience, prioritisation could account for both reward prediction errors and the policy's exploratory nature. This adjustment could help prioritise transitions where the policy was more exploratory (higher entropy), aligning more closely with SAC's exploration strategy.

\section{\textbf{Conclusion}}

    This paper introduces RPE-PER, a novel approach that enhances policy discovery by dynamically prioritising replay buffer experiences based on RPE. The predicted rewards are generated by EMCN, which offers a more comprehensive understanding of the environment by predicting rewards and next states compared to standard critic networks. RPE-PER integrates seamlessly with popular actor-critic RL frameworks such as TD3 and SAC, demonstrating effectiveness across diverse and complex environments, including high-dimensional action spaces, intricate balance tasks, and sparse rewards. This advancement in prioritised experience replay contributes to more efficient and practical Deep Reinforcement Learning for real-world continuous control scenarios.

\section*{Acknowledgements}
    The research was supported by Science For Technical Innovation (SfTI) on contract UoA3727019.
    Thanks to the Human Aspects in Software Engineering Lab (HASEL).
    
\bibliography{publication}

\begin{thebibliography}{}

\bibitem[\protect\citeauthoryear{Braun \bgroup \em et al.\egroup }{2018}]{braun2018retroactive}
Erin~Kendall Braun, G~Elliott Wimmer, and Daphna Shohamy.
\newblock Retroactive and graded prioritization of memory by reward.
\newblock {\em Nature communications}, 9(1):4886, 2018.

\bibitem[\protect\citeauthoryear{Brittain \bgroup \em et al.\egroup }{2019}]{brittain2019prioritized}
Marc Brittain, Josh Bertram, Xuxi Yang, and Peng Wei.
\newblock Prioritized sequence experience replay.
\newblock {\em arXiv preprint arXiv:1905.12726}, 2019.

\bibitem[\protect\citeauthoryear{Brockman \bgroup \em et al.\egroup }{2016}]{brockman2016openai}
Greg Brockman, Vicki Cheung, Ludwig Pettersson, Jonas Schneider, John Schulman, Jie Tang, and Wojciech Zaremba.
\newblock Openai gym.
\newblock {\em arXiv preprint arXiv:1606.01540}, 2016.

\bibitem[\protect\citeauthoryear{Fujimoto \bgroup \em et al.\egroup }{2018}]{fujimoto2018addressing}
Scott Fujimoto, Herke Hoof, and David Meger.
\newblock Addressing function approximation error in actor-critic methods.
\newblock In {\em International conference on machine learning}, pages 1587--1596. PMLR, 2018.

\bibitem[\protect\citeauthoryear{Fujimoto \bgroup \em et al.\egroup }{2020}]{fujimoto2020equivalence}
Scott Fujimoto, David Meger, and Doina Precup.
\newblock An equivalence between loss functions and non-uniform sampling in experience replay.
\newblock {\em Advances in Neural Information Processing Systems}, 33, 2020.

\bibitem[\protect\citeauthoryear{Ha and Schmidhuber}{2018}]{ha2018world}
David Ha and J{\"u}rgen Schmidhuber.
\newblock World models.
\newblock {\em arXiv preprint arXiv:1803.10122}, 2018.

\bibitem[\protect\citeauthoryear{Haarnoja \bgroup \em et al.\egroup }{2018}]{haarnoja2018soft}
Tuomas Haarnoja, Aurick Zhou, Pieter Abbeel, and Sergey Levine.
\newblock Soft actor-critic: Off-policy maximum entropy deep reinforcement learning with a stochastic actor.
\newblock In {\em International conference on machine learning}, pages 1861--1870. PMLR, 2018.

\bibitem[\protect\citeauthoryear{Henderson \bgroup \em et al.\egroup }{2018}]{henderson2018deep}
Peter Henderson, Riashat Islam, Philip Bachman, Joelle Pineau, Doina Precup, and David Meger.
\newblock Deep reinforcement learning that matters.
\newblock In {\em Proceedings of the AAAI conference on artificial intelligence}, volume~32, 2018.

\bibitem[\protect\citeauthoryear{Hessel \bgroup \em et al.\egroup }{2018}]{hessel2018rainbow}
Matteo Hessel, Joseph Modayil, Hado Van~Hasselt, Tom Schaul, Georg Ostrovski, Will Dabney, Dan Horgan, Bilal Piot, Mohammad Azar, and David Silver.
\newblock Rainbow: Combining improvements in deep reinforcement learning.
\newblock In {\em Proceedings of the AAAI conference on artificial intelligence}, volume~32, 2018.

\bibitem[\protect\citeauthoryear{Ibarz \bgroup \em et al.\egroup }{2021}]{ibarz2021train}
Julian Ibarz, Jie Tan, Chelsea Finn, Mrinal Kalakrishnan, Peter Pastor, and Sergey Levine.
\newblock How to train your robot with deep reinforcement learning: lessons we have learned.
\newblock {\em The International Journal of Robotics Research}, 40(4-5):698--721, 2021.

\bibitem[\protect\citeauthoryear{Januszewski}{2022}]{januszewski2022model}
Piotr Januszewski.
\newblock Model-free and model-based reinforcement learning, the intersection of learning and planning.
\newblock In {\em Proceedings of the 21st International Conference on Autonomous Agents and Multiagent Systems}, pages 1849--1851, 2022.

\bibitem[\protect\citeauthoryear{Lerner \bgroup \em et al.\egroup }{2021}]{lerner2021dopamine}
Talia~N Lerner, Ashley~L Holloway, and Jillian~L Seiler.
\newblock Dopamine, updated: reward prediction error and beyond.
\newblock {\em Current opinion in neurobiology}, 67:123--130, 2021.

\bibitem[\protect\citeauthoryear{Lin}{1992}]{lin1992self}
Long-Ji Lin.
\newblock Self-improving reactive agents based on reinforcement learning, planning and teaching.
\newblock {\em Machine learning}, 8:293--321, 1992.

\bibitem[\protect\citeauthoryear{Mnih \bgroup \em et al.\egroup }{2015}]{mnih2015human}
Volodymyr Mnih, Koray Kavukcuoglu, David Silver, Andrei~A Rusu, Joel Veness, Marc~G Bellemare, Alex Graves, Martin Riedmiller, Andreas~K Fidjeland, Georg Ostrovski, et~al.
\newblock Human-level control through deep reinforcement learning.
\newblock {\em nature}, 518(7540):529--533, 2015.

\bibitem[\protect\citeauthoryear{Oh \bgroup \em et al.\egroup }{2021}]{oh2021model}
Youngmin Oh, Jinwoo Shin, Eunho Yang, and Sung~Ju Hwang.
\newblock Model-augmented prioritized experience replay.
\newblock In {\em International Conference on Learning Representations}, 2021.

\bibitem[\protect\citeauthoryear{Pathak \bgroup \em et al.\egroup }{2017}]{pathak2017curiosity}
Deepak Pathak, Pulkit Agrawal, Alexei~A Efros, and Trevor Darrell.
\newblock Curiosity-driven exploration by self-supervised prediction.
\newblock In {\em International conference on machine learning}, pages 2778--2787. PMLR, 2017.

\bibitem[\protect\citeauthoryear{Rouhani \bgroup \em et al.\egroup }{2024}]{rouhani2024building}
Nina Rouhani, David Clewett, and James~W Antony.
\newblock Building and breaking the chain: A model of reward prediction error integration and segmentation of memory.
\newblock {\em Journal of Cognitive Neuroscience}, pages 1--13, 2024.

\bibitem[\protect\citeauthoryear{Saglam \bgroup \em et al.\egroup }{2022}]{saglam2022actor}
Baturay Saglam, Furkan~B Mutlu, Dogan~C Cicek, and Suleyman~S Kozat.
\newblock Actor prioritized experience replay.
\newblock {\em arXiv preprint arXiv:2209.00532}, 2022.

\bibitem[\protect\citeauthoryear{Schaul \bgroup \em et al.\egroup }{2015}]{schaul2015prioritized}
Tom Schaul, John Quan, Ioannis Antonoglou, and David Silver.
\newblock Prioritized experience replay.
\newblock {\em arXiv preprint arXiv:1511.05952}, 2015.

\bibitem[\protect\citeauthoryear{Schultz}{2017}]{schultz2017reward}
Wolfram Schultz.
\newblock Reward prediction error.
\newblock {\em Current Biology}, 27(10):R369--R371, 2017.

\bibitem[\protect\citeauthoryear{Shelhamer \bgroup \em et al.\egroup }{2016}]{shelhamer2016loss}
Evan Shelhamer, Parsa Mahmoudieh, Max Argus, and Trevor Darrell.
\newblock Loss is its own reward: Self-supervision for reinforcement learning.
\newblock {\em arXiv preprint arXiv:1612.07307}, 2016.

\bibitem[\protect\citeauthoryear{Simmons-Edler \bgroup \em et al.\egroup }{2019}]{simmons2019reward}
Riley Simmons-Edler, Ben Eisner, Daniel Yang, Anthony Bisulco, Eric Mitchell, Sebastian Seung, and Daniel Lee.
\newblock Reward prediction error as an exploration objective in deep rl.
\newblock {\em arXiv preprint arXiv:1906.08189}, 2019.

\bibitem[\protect\citeauthoryear{Todorov \bgroup \em et al.\egroup }{2012}]{todorov2012mujoco}
Emanuel Todorov, Tom Erez, and Yuval Tassa.
\newblock Mujoco: A physics engine for model-based control.
\newblock In {\em 2012 IEEE/RSJ international conference on intelligent robots and systems}, pages 5026--5033. IEEE, 2012.

\bibitem[\protect\citeauthoryear{Zha \bgroup \em et al.\egroup }{2019}]{zha2019experience}
Daochen Zha, Kwei-Herng Lai, Kaixiong Zhou, and Xia Hu.
\newblock Experience replay optimization.
\newblock {\em arXiv preprint arXiv:1906.08387}, 2019.

\bibitem[\protect\citeauthoryear{Zhou \bgroup \em et al.\egroup }{2022}]{zhou2022continuously}
Zihan Zhou, Wei Fu, Bingliang Zhang, and Yi~Wu.
\newblock Continuously discovering novel strategies via reward-switching policy optimization.
\newblock {\em arXiv preprint arXiv:2204.02246}, 2022.

\end{thebibliography}
\bibliographystyle{named}
\end{document}